\pdfoutput=1

\documentclass[11pt]{article}

\usepackage[]{acl}

\usepackage{times}
\usepackage{latexsym}

\usepackage[T1]{fontenc}

\usepackage[utf8]{inputenc}

\usepackage{microtype}

\usepackage{booktabs}
\usepackage{times}
\usepackage{latexsym}
\usepackage{amsmath}
\usepackage{amssymb}
\usepackage{subcaption}
\usepackage{float}
\usepackage{multirow}
\usepackage{graphicx}
\usepackage{listings}
\usepackage{adjustbox}

\usepackage{amsthm}
\newtheorem{definition}{Definition}

\usepackage{pifont}

\usepackage{xspace}
\usepackage[capitalise]{cleveref}
\usepackage{paralist}

\makeatletter
\newcommand*{\etc}{%
    \@ifnextchar{.}%
        {etc}%
        {etc.\@\xspace}%
}
\makeatother

\usepackage{xcolor}

\title{Few-Shot Cross-lingual Transfer for Coarse-grained \\ De-identification of Code-Mixed Clinical Texts}

\author{
  Saadullah Amin$^{1,2}$, Noon Pokaratsiri Goldstein$^{1,2}$, Morgan K. Wixted$^{1,2}$\\
  \textbf{Alejandro García-Rudolph}$^3$\textbf{,} \textbf{Catalina Martínez-Costa}$^4$\textbf{,} \textbf{G\"unter Neumann}$^{1,2}$\\
  $^1$Department of LST, Saarland University, Germany\quad $^2$DFKI, SIC, Germany \\ 
  $^3$Department of Research and Innovation, Institut Guttmann, UAB, Spain\\
  $^4$Department Informática y Sistemas, Universidad de Murcia, IMIB-Arrixaca, Spain\\
  \small\texttt{\{saadullah.amin,noon.pokaratsiri,morgan.wixted,guenter.neumann\}@dfki.de}\\ 
  \small\texttt{agarciar@guttmann.com\quad cmartinezcosta@um.es}
}

\begin{document}

\maketitle


\begin{abstract}
Despite the advances in digital healthcare systems offering curated structured knowledge, much of the critical information still lies in large volumes of unlabeled and unstructured clinical texts.
These texts, which often contain protected health information (PHI), are exposed to information extraction tools for downstream applications, risking patient identification.
Existing works in de-identification rely on using large-scale annotated corpora in English, which often are not suitable in real-world multilingual settings.
Pre-trained language models (LM) have shown great potential for \emph{cross-lingual transfer} in low-resource settings. 
In this work, we empirically show the \emph{few-shot cross-lingual transfer} property of LMs for named entity recognition (NER) and apply it to solve a low-resource and real-world challenge of code-mixed (Spanish-Catalan) clinical notes de-identification in the stroke domain.
We annotate a gold evaluation dataset to assess few-shot setting performance where we only use a few hundred labeled examples for training.
%
Our model improves the zero-shot F1-score from 73.7\% to 91.2\% on the gold evaluation set when adapting Multilingual BERT (mBERT) \cite{devlin2019bert} from the MEDDOCAN \cite{marimon2019automatic} corpus with our few-shot cross-lingual target corpus.
When generalized to an out-of-sample test set, the best model achieves a human-evaluation F1-score of 97.2\%.
\end{abstract}



\section{Introduction} \label{sec:introduction}
With growing interest and innovations in data-driven digital technologies, privacy has become an important legal topic for the technology to be regulations-compliant.
In Europe, the General Data Protection Regulation (GDPR) \cite{regulation2016regulation} requires data owners to have a legal basis for processing personally identifiable information (PII), which also includes the explicit consent of the subjects.
In cases where explicit consent is not possible, anonymization is often seen as a resorted-to solution.
Clinical texts contain rich information about patients, including their gender, age, profession, residence, family, and history, that is useful for record keeping and billing purposes \cite{johnson2016mimic,shickel2017deep}. 
%


\begin{figure}[!t]
    \includegraphics[width=0.85\linewidth]{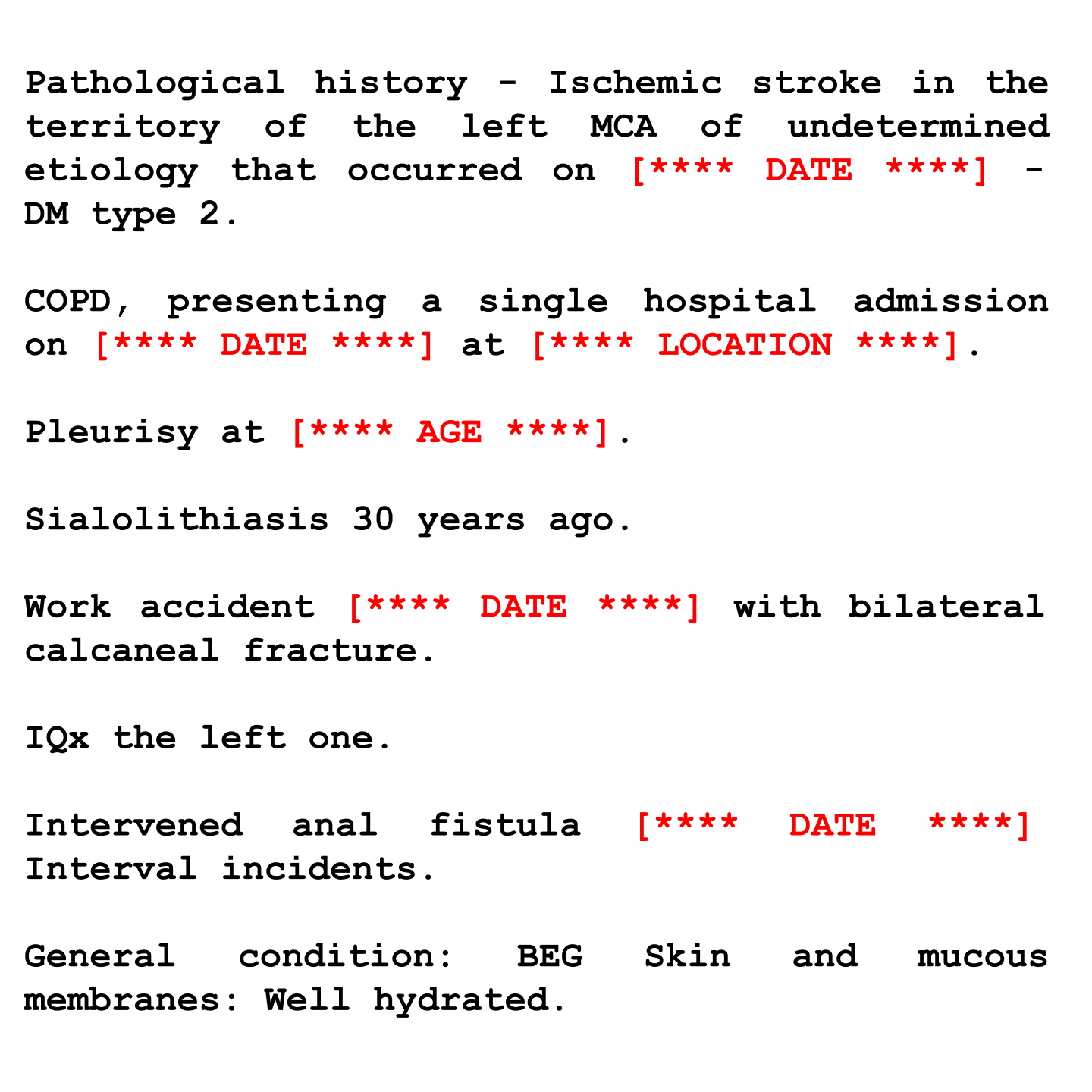}
    \centering
    \caption{
        The process of text de-identification involves the removal of a predefined set of direct identifiers in text \cite{elliot2016anonymisation}. 
        For clinical notes, this set is often the PHI categories (or types) defined by the Health Insurance Portability and Accountability Act (HIPAA) \cite{gunn2004health}.
        The example here shows a de-identified excerpt of a patient note from the Spanish-Catalan stroke dataset used in this study (translated into English here for readability).
    }
    \label{fig:anon_example}
\end{figure}


%
In this work, we focus on the task of removing PHI from clinical texts, also called de-identification (\cref{fig:anon_example}).
We address a real-world challenge where the target texts are code-mixed (Spanish-Catalan) and domain-constrained (stroke).
To avoid high annotation costs, we consider a more realistic setting where we annotate a gold evaluation corpus and a few hundred examples for training.
Our approach is motivated by strong performance of pre-trained LMs in few-shot cross-lingual transfer for NER with high sample efficiency (see \cref{fig:few_shot_conll_i2b2}) in comparison to supervised or unsupervised approaches.
Our contributions are summarized as follows:
\begin{itemize}
    \item We empirically show the few-shot cross-lingual transfer property of multi-lingual pre-trained LM, mBERT, for NER.
	\item We apply this property to solve a low-resource problem of code-mixed and domain-specific clinical note de-identification.
	\item We annotate a few-shot training corpus and a gold evaluation set, minimizing annotation costs while achieving a significant performance boost without needing a large-scale labeled training corpus.
\end{itemize}
%


\section{Related Work}
GDPR-compliant anonymization requires complete and irreversible removal of any information that may lead to a subject's data being identified (directly or indirectly) from a dataset \cite{elliot2016anonymisation}.
However, de-identification is limited to removing specific predefined direct identifiers; further replacement of such direct identifiers with pseudonyms is referred to as pseudonymization \cite{alfalahi2012pseudonymisation}.
Generally, de-identification can be seen as a subset of anonymization despite interchangeable usage of the terms in the literature \cite{chevrier2019use}.
We focus on solving the problem of de-identification in the clinical domain as a sequence labeling task, specifically named entity recognition (NER) \cite{lample2016neural}.
%


\subsection{Clinical De-identification}
2014 i2b2/UTHealth \cite{stubbs2015annotating}, and the 2016 CEGS N-GRID \cite{stubbs2017identification} shared tasks explore the challenges of clinical de-identification on diabetic patient records and psychiatric intake records respectively.
Earlier works include machine learning and rule-based approaches \cite{meystre2010automatic,yogarajan2018survey}, with \citet{liu2017identification} and \citet{dernoncourt2017identification} being the first to propose neural architectures.
\citet{friedrich2019adversarial} propose an adversarial approach to learn privacy-preserving text representations; \citet{yang2019study} use domain-specific embeddings trained on unlabeled corpora.
While most works have mainly focused on English, some efforts have been made for Swedish \cite{velupillai2009developing,alfalahi2012pseudonymisation} and Spanish (with a synthetic dataset at the MEDDOCAN shared task \cite{marimon2019automatic}).
As outlined in \citet{lison2021anonymisation}, a significant challenge in clinical text de-identification is the lack of labeled data.
\citet{hartman2020customization} show that a small number of manually labeled PHI examples can significantly improve performance.
Prior works in few-shot NER consider the problem where a model is trained on one or more source domains and tested on unseen domains with a few labeled examples per class, some of which with entity tags different from those in the source domains \cite{yang2020simple}.
%
Models are trained with prototypical methods, noisy supervised pre-training, or self-labeling \cite{huang2020few}.
We consider a setting where the target and source domains share the \emph{same} entity (PHI) tags, but with a few labeled examples in the target domain (or language). 
%
A similar setup has been employed in few-shot question answering \cite{ram2021few}.
%


\section{Problem Statement} \label{sec:methodology}
We approach the de-identification problem as an NER task.
Given an input sentence $\mathbf{x}$ with $N$ words: $\mathbf{x}=[x_i]_{i=1:N}$, we feed it to an encoder $f_{\phi}: \mathbb{R}^{N} \rightarrow \mathbb{R}^{N \times d}$ to obtain a sequence of hidden representations $\mathbf{h}=[h_i]_{i=1:N}$
\begin{align*}
\mathbf{h} = f_{\phi}(\mathbf{x}).
\end{align*}
We feed $\mathbf{h}$ into the NER classifier which is a linear classification layer with the \texttt{softmax} activation function to predict the PHI label of $\mathbf{x}$:
\begin{align*}
p_{\theta}(\mathbf{Y}|\mathbf{x}) = \texttt{softmax}(\mathbf{W}^T\mathbf{h} + \mathbf{b}).
\end{align*}
$p_{\theta}(\mathbf{Y}|\mathbf{x}) \in \mathbb{R}^{N \times |\mathcal{P}|}$ is the probability distribution of PHI labels for sentence $\mathbf{x}$ and $\mathcal{P}$ is the PHI label set. 
$\theta = [\phi, \mathbf{W} \in \mathbb{R}^{d \times |\mathcal{P}|}, \mathbf{b} \in \mathbb{R}^{|\mathcal{P}|}]$ denote the set of learnable parameters and $d$ being the hidden dimension.
The model is trained to minimize the per-sample negative log-likelihood: 
\begin{align}
\mathcal{L} = -\frac{1}{N} \sum_{i=1}^{N} \log p_{\theta}(Y_i=y_i|x_i).
\end{align}
For pre-trained LMs, this setting corresponds to NER fine-tuning \cite{wu2019beto}. When we jointly fine-tune on more than one NER dataset, we refer to it as multi-task learning.

\begin{definition}[\textbf{Few-Shot NER}]\label{def:1}
Given an entity label set $\mathcal{P}$, we define the task of few-shot NER as having access to $K\leq M$ labeled sentences containing each element $ p \in \mathcal{P}$ at least once, where $K$ is a small number (e.g., in $[50,500]$) and $M$ is orders of magnitude larger (e.g., $\geq 1000$).
\end{definition}


\begin{figure*}[!t]
    \includegraphics[width=1.0\linewidth]{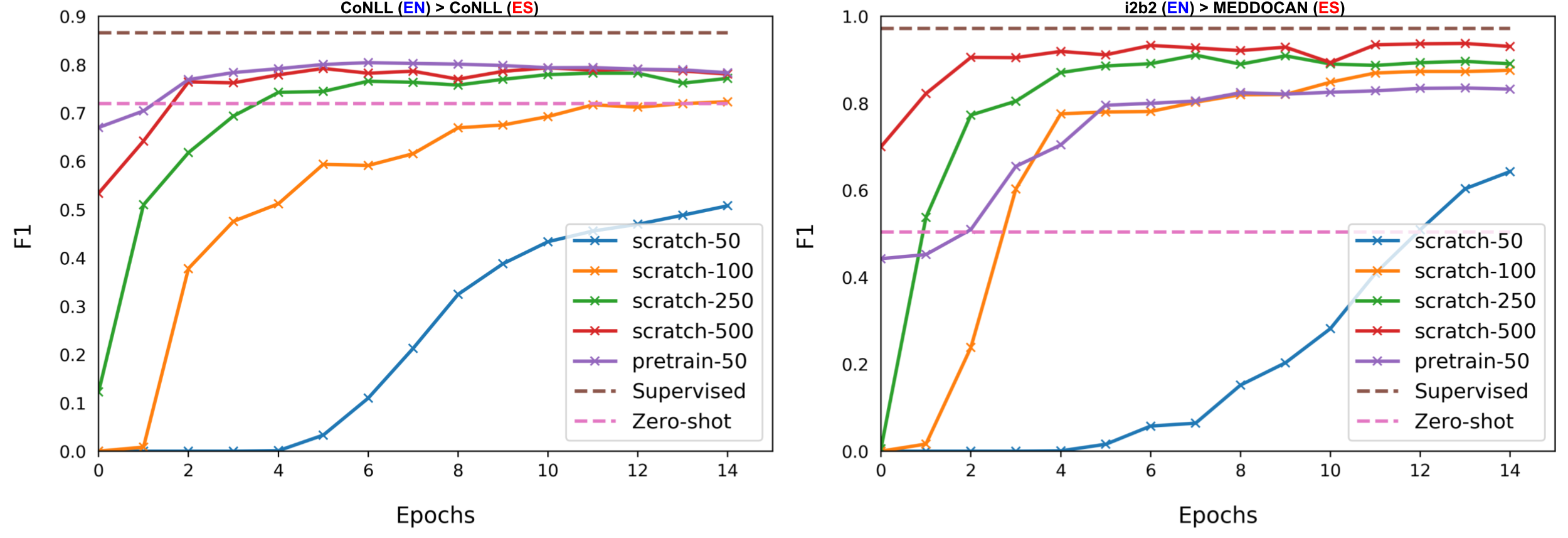}
    \centering
    \caption{
        \textbf{Few-shot cross-lingual NER transfer of mBERT}. We compare different transfer learning scenarios from English (EN) to Spanish (ES) for two pairs of datasets as preliminary study to investigate the effectiveness of the few-shot cross-lingual transfer in mBERT: CoNLL-2003 to CoNLL-2002 (\emph{left}) and i2b2-2014 to MEDDOCAN (\emph{right}).
        We use supervised fine-tuning on the full training set of the target language (ES) as the \emph{upper bound} and the zero-shot score of the model
        (pre-trained only on the source language (EN) training set) as the \emph{lower bound} for target language (ES) performance.
        We then consider 50, 100, 250, and 500 examples from the target language as few-shot training corpora and train the models for 15 epochs.
        The models without any pre-training on the source corpus (scratch) eventually outperform the \emph{lower bound} as the number of examples grow; with sufficient epochs, the model with only 50 target-language samples reaches more than 10\% gain in de-identification task (\emph{right: scratch-50}).
        However, we find the cross-lingual transfer-learning strategy to be most sample efficient when pre-trained on \emph{source} language---50-shot performance (\emph{left \& right: pretrain-50}) comparable to 500-shot (\emph{left \& right: scratch-500}). We apply this strategy to address the real-world challenges of Spanish-Catalan de-identification.
    }\label{fig:few_shot_conll_i2b2}
\end{figure*}


\begin{definition}[\textbf{Few-Shot NER Transfer}]\label{def:2}
Given an NER dataset in a source domain (or language), we define the task of few-shot cross-domain (or cross-lingual) NER transfer as adapting a model trained on the source domain (or language) to a target domain (or language) with access to a few-shot corpus (\cref{def:1}). 
\end{definition}

\noindent This setting is different from prior studies in NER transfer including few-shot \cite{huang2020few}, unsupervised \cite{keung2020don}, and semi-supervised NER \cite{amin2021t2ner}.


\subsection{Few-Shot Cross-Lingual Transfer}
mBERT \cite{devlin2019bert} has been shown to achieve strong performance for zero-shot cross-lingual transfer tasks, including NER \cite{wu2019beto, pires2019multilingual}. 
Adversarial learning has been applied with limited gains \cite{keung2019adversarial} in unsupervised approaches to improve zero-shot NER transfer, whereas feature alignments have shown better results \cite{wang2019cross}.
%
Meta-learning with minimal resources \cite{wu2020enhanced} and word-to-word translation \cite{wu2021unitrans} have shown further performance gains. 
The current state-of-the-art approach \cite{chen2021advpicker} combines token-level adversarial learning with self-labeled data selection and knowledge distillation.
%

 
\begin{table}
    \centering
    \resizebox{5cm}{!}{
    \begin{tabular}{cc}
        \toprule
        \textsc{CoNLL (EN)} $\rightarrow$ \textsc{CoNLL (ES)} & \textsc{F1} \\
        \midrule
        \citet{pires2019multilingual} & 73.59 \\
        \citet{wu2019beto} & 74.96 \\
        \citet{keung2019adversarial} & 74.30 \\
        \citet{wang2019cross} & 75.77 \\
        \citet{wu2020enhanced} & 77.30 \\
        \citet{wu2021unitrans} & 76.75 \\
        \citet{chen2021advpicker} & \textbf{79.00} \\
        \midrule
        \textit{few-50 (or pretrain-50)} & 78.30 \\
        \bottomrule 
    \end{tabular}
    }
    \caption{
        Cross-lingual transfer results on CoNLL. \textit{few-50} represents our fine-tuning of EN trained mBERT with 50 random labeled samples from ES.
    }
    \label{table:few_en_es_transfer}
 \end{table}
 
\begin{figure}[!t]
\includegraphics[width=1.0\linewidth]{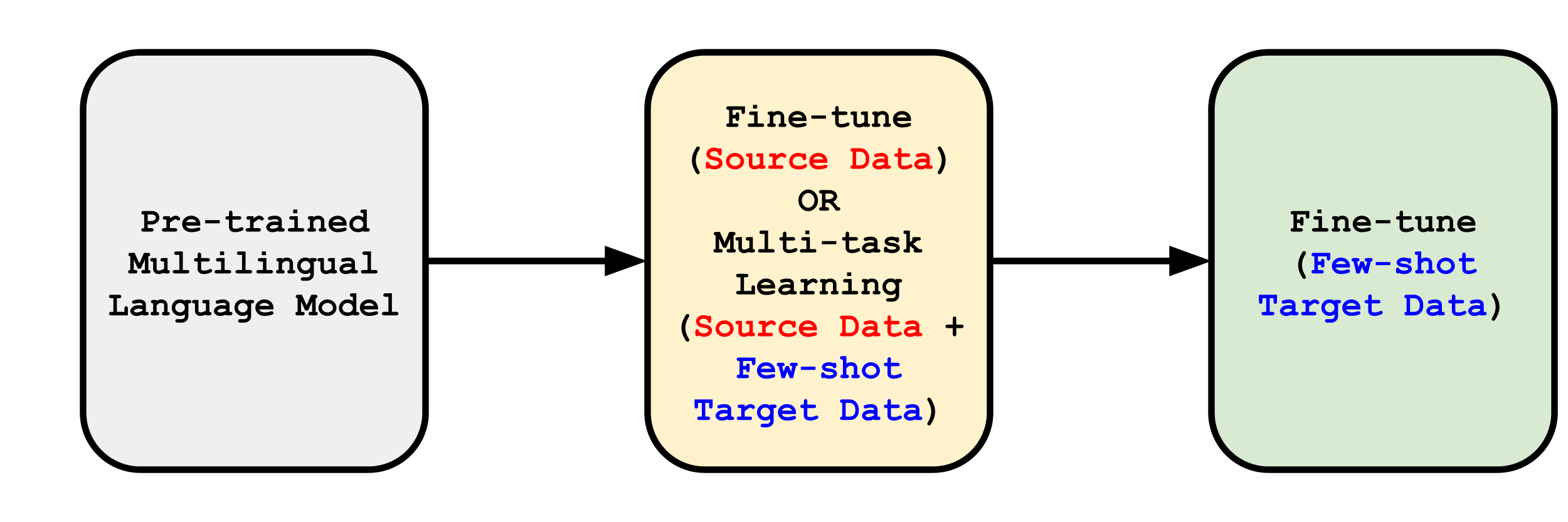}
    \centering
    \caption{
        Our \textit{few-shot cross-lingual transfer} strategy for clinical text de-identification.
    }
    \label{fig:transfer_strategy}
\end{figure}

%
To investigate the few-shot transferability of mBERT, we consider two pairs of datasets with English as the source language and Spanish as the target language: the CoNLL-2003/CoNLL-2002 \cite{tjong-kim-sang-de-meulder-2003-introduction, tjong-kim-sang-2002-introduction} in the \emph{general domain} and i2b2/MEDDOCAN \cite{stubbs2015annotating, marimon2019automatic} in the \emph{clinical domain}.
We report results of our preliminary study in \cref{fig:few_shot_conll_i2b2}. We observed that with as few as 50 random labeled training samples from the target language, we obtain substantial gains for both datasets, with near state-of-the-art on CoNLL (\cref{table:few_en_es_transfer}).
We refer to this as \emph{few-shot cross-lingual transfer} property of mBERT for NER.
Our study highlights that the property holds for different domains (general and clinical), where the latter focuses on the de-identification task.
We leave a large-scale study on more datasets with different languages and domains as future work.  
Compared to supervised (unsupervised) methods, which use complete labeled (unlabeled) target data, our few-shot approach is \emph{sample-efficient} and alleviates the need of complex pipelines \cite{wu2020enhanced,wu2021unitrans,chen2021advpicker} and large-scale annotations.
Furthermore, \citet{keung2020don} highlights the spurious effects of using source data as a development set and recommends using target data as a development set for model selection in NER transfer.
Our findings and those in \citet{hartman2020customization} motivate us to \emph{(a)} propose an optimal \emph{few-shot cross-lingual transfer} strategy (outlined in \cref{fig:transfer_strategy}), \emph{(b)} annotate a target development set, and \emph{(c)} construct an annotated \emph{few-shot target corpus} for effective cross-lingual transfer learning.
%


\section{Data and Annotation} \label{sec:guttmann_data}
Our dataset consists of stroke patient records collected at Instiut Guttmann.\footnote{\url{https://www.guttmann.com/ca/institut-universitari-guttmann-uab}} \cref{table:raw_data_stats} summarizes the raw data statistics and \cref{table:clinical_topics} in \cref{app:data_desc} describes the topics present in the texts.
%
%
%
We set aside 100 randomly sampled notes for out-of-sample generalizability evaluation and consider the remaining notes for our development and few-shot corpora sampling; 396$k$ sentences are tokenized in the process.
%
%

%
Following the protocol in \citet{gao2021manual} for constructing manually annotated distantly supervised relation extraction test sets, we train mBERT on the MEDDOCAN corpus, using coarse-grained PHI categories $\{$DATE, AGE, LOCATION, NAME, CONTACT, PROFESSION, ID$\}$ with the BIO scheme \cite{farber2008improving}, for evaluation and few-shot training data selection.
%
%
We use the trained model to make predictions on the dataset and observe that the model predicts PHI on only 50$k$ out of the 396$k$ sentences. 
%
%
A dataset of 5000 sentences ($<$ 2\% of raw sentences) is constructed from a mix of randomly sampled 2500 sentences from this 50$k$ and 2500 from the remaining sentences.
%
%
We split the dataset into two partitions of 2500 sentences for independent annotation by two annotators.
The annotation is performed one sentence at a time by applying one of the 7 coarse-grained PHI labels to each token using the \textsc{T2NER-annotate} toolkit \cite{amin2021t2ner}.
%
Each annotator's confidence level between 1-5 is recorded for the token-level labels for each sample.
To record the inter-annotator agreement, we use token-level Cohen's kappa \cite{cohen1960coefficient} statistic reaching a value of 0.898.
In total, the two annotators agreed on 3924 sentences, resulting in our final evaluation set. 
To save annotation costs for developing a \textit{few-shot target corpus}, we resolved the disagreements to obtain a 384-sentence few-shot corpus for training (see \cref{app:data_anno} for annotation details).
%
%
%
%

%
Our source dataset (the MEDDOCAN corpus) consists of 1000 synthetically generated clinical case studies in Spanish \cite{marimon2019automatic}. 
%
The corpus was selected manually by a practicing physician and augmented with PHI from discharge summaries and clinical records. 
%
In contrast, our target corpus focuses on the stroke domain and contains PHI from real-world records.
%
%
Since the target data is code-mixed between Spanish and Catalan, with the majority (53\%) being Catalan, the transfer from Spanish source data (MEDDOCAN) is cross-lingual.\footnote{Although similar, Spanish and Catalan are distinct languages. The domain of MEDDOCAN is missing an explicit mention in \citet{marimon2019automatic}.} 
%


\begin{table}[!t]
    \centering
    \resizebox{\linewidth}{!}{
    \begin{tabular}{ccccc}
        \toprule
        \textbf{Patients} & \textbf{Notes} & \textbf{ES} & \textbf{CA} & \textbf{Other}  \\
        \midrule
        1,500 & 327,775 & 42.8\% & 53.0\% & 4.2\% \\
        \bottomrule
    \end{tabular}
    }
    \caption{
        Raw statistics of the Spanish (ES)-Catalan (CA) data from stroke domain. 
    }
    \label{table:raw_data_stats}
 \end{table}



\begin{table*}[!t]
    \centering
    \resizebox{10cm}{!}{
        \begin{tabular}{lccc}
            \toprule
            \textbf{Transfer Strategy} & \textbf{Precision} & \textbf{Recall} & \textbf{F1} \\
            \midrule
            \textsc{Fine-tune} (\texttt{M}) & 80.1 & 68.2 & 73.7 \\
            \textsc{Fine-tune} (\texttt{M}) $\rightarrow$ \textsc{Fine-tune} (\texttt{F}) & 83.5 & 94.2 & 88.6 \\
            \midrule
            \textsc{Multi-task} (\texttt{M} + \texttt{F}) & 86.0 & 93.3 & 89.5 \\
            \textsc{Multi-task} (\texttt{M} + \texttt{F}) $\rightarrow$ \textsc{Fine-tune} (\texttt{F}) & \textbf{87.7} & \textbf{95.0} & \textbf{91.2} \\
            \bottomrule
        \end{tabular}
    }
    \caption{
        Results on the development set from the code-mixed stroke data. 
        \texttt{M} denotes the MEDDOCAN \cite{marimon2019automatic} training set (source) normalized to 7 PHIs (see \cref{app:meddocan_normalization}) at sentence level and \texttt{F} denotes our few-shot target corpus. Here multi-task learning refers to the joint fine-tuning on two datasets.
    }
    \label{table:main_results}
\end{table*}



\section{Experiments and Results}
We conduct our experiments with the \textsc{T2NER} framework \cite{amin2021t2ner}.\footnote{\url{https://github.com/suamin/T2NER}}
%
%
For the baseline, we consider zero-shot performance on the evaluation set of the mBERT encoder fine-tuned on the MEDDOCAN training set consisting of 16,299 samples.
We then fine-tune it on the few-shot target corpus as outlined in \cref{fig:transfer_strategy}.
Following the multi-task learning \cite{lin2018multi} approach in \textsc{T2NER}, we jointly fine-tune mBERT on the MEDDOCAN and few-shot target corpora.
Since the few-shot corpus is much smaller, the multi-task learning helps the model transfer. It further acts as a regularization approach by sharing parameters between the datasets.
To improve performance on the target data, we further fine-tune with the few-shot target corpus after the first step of fine-tuning to have improved target performance; for the model to be an expert in target \cite{cao2020unsupervised}. 
%
%
All the models are trained for 3 epochs with a learning rate of 3e-5 and linear warm-up of 10\%.
For few-shot fine-tuning only, the model is trained for 25 epochs.
%
%

\begin{figure}[!t]
    \includegraphics[width=1.0\linewidth]{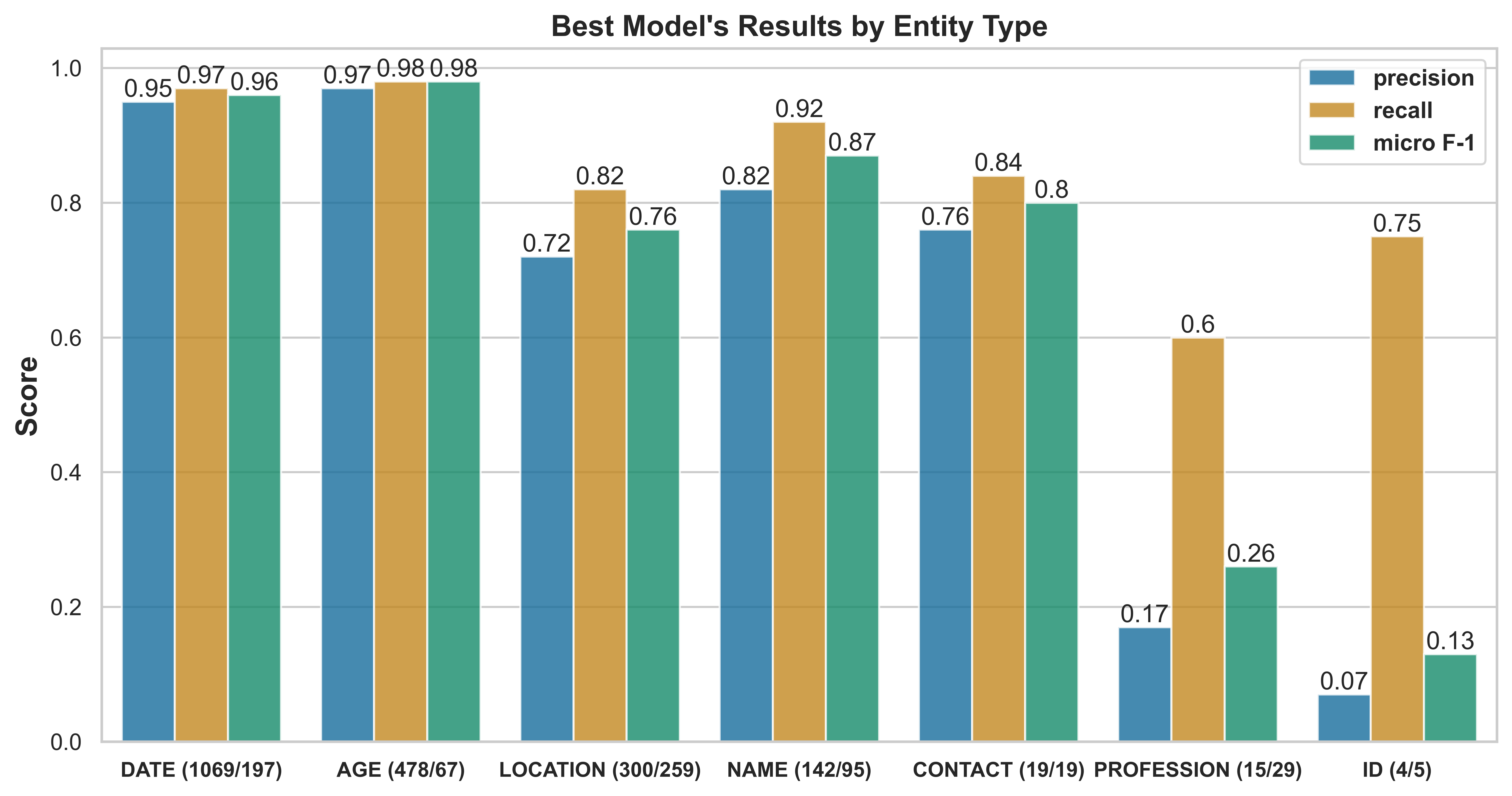}
    \centering
    \caption{NER metrics on the evaluation set for each entity type with their frequency (dev/few-train).}
    \label{fig:barplot}
\end{figure}

\cref{table:main_results} shows our results. 
Fine-tuning the baseline mBERT model with the few-shot target corpus improves the F1-score from 73.7\% to 88.6\%, a substantial gain of 14.9\%, highlighting the effectiveness of \emph{few-shot cross-lingual transfer} with mBERT.
%
%
The significant increase in recall (26\% points) compared to precision (3.4\% points) suggests an increase in the model's capacity to recognize domain-specific entities.
%
%
%
Multi-task fine-tuning improves the F1-score to 89.5\%; further fine-tuning on the few-shot target corpus boosts the best model's performance to 91.2\%. 
\cref{fig:barplot} shows per-PHI-label scores on the development set, along with their frequency. 
The model performs almost perfectly on DATE and AGE, since most DATE and AGE labeled segments are similar between Spanish and Catalan as they are simple numbers (for DATE) and numbers followed by the word (for AGE; `\emph{edad}' in both Spanish and Catalan). 
There are some differences in time expressions, e.g., day of the week, as the words are distinctly dissimilar. 
However, structurally there is only a slight difference.
Further, the model struggles with the ID class due to low sample size (5 instances in the few-shot corpus), and it is generally challenging to disambiguate between an alphanumeric string and a PHI ID, as also noted by the ID class' high recall.
Our error analysis reveals high false positives for the PROFESSION label in Catalan, e.g.: `\emph{Coloma de Gramenet}' (a LOCATION) and `\emph{Dialogant}' (being able to communicate) are both labeled as PROFESSION.
To test the model's generalizability, we tokenize the 100 out-of-sample notes into sentences and make predictions with our best model.
The resulting annotated sentences are reconstructed into patient notes, which are manually evaluated by two reviewers (one external and one annotator) for occurrences of true and false positives and negatives.
%
%
The model achieves precision, recall, and F1-scores of 95.1\%, 99.3\%, and 97.1\% respectively on the out-of-sample notes, highlighting the effectiveness of our approach. 
%


\section{Conclusion}
We address the task of clinical notes de-identification in a low-resource scenario. 
%
%
By investigating the \emph{few-shot cross-lingual transfer} property of mBERT, we propose a strategy that significantly boosts zero-shot performance 
while keeping the number of required annotated samples low. 
%
Our results highlight the effectiveness of the proposed strategy for the task 
with a potential for future applications in other low-resource scenarios.
%
%



\section*{Acknowledgments}
The authors would like to thank the anonymous reviewers and Josef van Genabith for their helpful feedback.
The work was partially funded by the European Union (EU) Horizon 2020 research and innovation programme through the project Precise4Q (777107), the German Federal Ministry of Education and Research (BMBF) through the project CoRA4NLP (01IW20010), the grant RTI2018-099039-J-I00 funded by MCIN/AEI/10.13039/5011000110033/ and “FEDER”, by the EU.
The authors also acknowledge the cluster compute resources provided by the DFKI.
%


\bibliography{custom}
\bibliographystyle{acl_natbib}

\appendix

\setcounter{table}{0}
\renewcommand{\thetable}{A.\arabic{table}}

\setcounter{figure}{0}
\renewcommand{\thefigure}{A.\arabic{figure}}


\section{Additional Dataset Details} \label{app:data_desc}
In addition to the 7 coarse-grained PHI entities in \cite{stubbs2015annotating}, our dataset contains cross-sentence recurring entities about topics that may be of interest in the clinical domain. 
These topics are grouped by their potential clinical application areas and are summarized in \cref{table:clinical_topics}. 
The label frequency distribution, as noted in \cref{fig:barplot}, is consistent with general characteristics of medical notes, which usually highlight notable events such as symptom onsets, procedures, admissions, transfers, and discharges, in addition to the date of each documentation. 
As a result, they tend to contain a higher frequency for the DATE PHI.
In addition, the lower occurrence of the NAME PHI compared to the AGE and LOCATION entities is consistent with how healthcare providers usually communicate patient's information. 
Healthcare providers are trained to refer to patients simply by their age, gender, and the appropriate diagnosis to avoid inadvertently sharing HIPAA-sensitive information, e.g., \emph{"a 60-year-old male with ischemic stroke admitted on [DATE] from [LOCATION] (...)"}. 
The patient's name may be used at the beginning of a medical note; however, subsequent anaphoric references are often accomplished via pronouns, omitting the NAME entity in the process. 
In addition, as it is applicable to Spanish medical records, nominative pronouns anaphorically referencing a patient may be omitted as they are grammatical in Spanish. 
We avoid releasing our dataset due to presence of real PHI information. 
We will consider replacing the real PHI with synthetic ones, similar to MEDDOCAN, for a GDPR-compliant release.
%



\begin{table}[!t]
\centering
\begin{adjustbox}{width=0.98\linewidth}
\begin{tabular}{@{}ll@{}}
\toprule
\textbf{Topic Areas} & \textbf{Subcategories} \\ \midrule
      \begin{tabular}[c]{@{}l@{}}Diagnostics \& \\ 
      Treatments
      \end{tabular}
      & 
      \begin{tabular}[c]{@{}l@{}}Ischemic vs Hemorrhagic \\ 
      Affected areas and vessels \\
      Comorbidities \\
      Medication history \\
      Associated lifestyle factors \\
      Treatments and interventions
      \end{tabular} \\ \midrule
      \begin{tabular}[c]{@{}l@{}}Symptoms \& \\ 
      Monitoring
      \end{tabular}
      & 
      \begin{tabular}[c]{@{}l@{}}Vital signs \\
      Lab results and cultures \\
      Pain and comfort \\
      Bladder and bowel controls  
      \end{tabular} \\ \midrule
      \begin{tabular}[c]{@{}l@{}}
      Long-term Care \& \\ 
      Discharge Planning
      \end{tabular}
      &
      \begin{tabular}[c]{@{}l@{}}Mobility \\
      Cognitive ability \\
      Nutrition \\
      Psychosocial factors
      \end{tabular} \\
      \bottomrule
\end{tabular}
\end{adjustbox}
\caption{Topics and subcategories in the corpus.}
\label{table:clinical_topics}
\end{table}



\section{Annotation} \label{app:data_anno}


\subsection{Annotator Profile} \label{app:anno_profile}
Two graduate research assistants completed the annotation of the dataset. 
Both annotators have at a minimum CEFR\footnote{\url{https://www.coe.int/en/}} B2-C1 Spanish (Castilian) proficiency.
One annotator also has clinical experience in the cardiovascular and cerebrovascular specialty, including knowledge of Spanish medical terminology.
Neither annotator has formal training in Catalan; both have prior experience working with text data in the language in this domain.
%



\subsection{Annotation Guidelines} \label{app:anno_guide}
The annotation process followed criteria for each entity as described in \citet{stubbs2015annotating}. 
The 7 entities: AGE, CONTACT, DATE, ID, LOCATION, NAME, and PROFESSION represent a larger granularity of the 18 HIPAA-defined PHI \cite{stubbs2015annotating}.
We examined the training sets of i2b2 \cite{stubbs2015annotating} and MEDDOCAN \cite{marimon2019automatic} and adapted the i2b2 annotation guidelines to create our own annotation guidelines.
This step was necessary since we only focused on coarse-grained PHI types compared to fine-grained types considered in these two datasets.
%
The adjusted guidelines utilized in this annotation process are summarized in \cref{table:anno_guidelines}.
%



\begin{table}[!t]
    \centering
    \begin{adjustbox}{width=0.98\linewidth}
        \begin{tabular}{@{}lcllll@{}}
            \toprule
            \multicolumn{1}{l}{\textbf{Description}} & \textbf{Observation} \\ 
            \midrule
            \ \ Sentences annotated by annotator (A) & 4400 \\
            \ \ Sentences annotated by annotator (B) & 4343 \\
            \midrule
            \ \ Sentences annotated and revised by (A, B) & \textbf{4314} \\
            \ \ \ \ Agreements & 3924 \\ 
            \ \ \ \ Disagreements & 390 \\
            \ \ \ \ Token-level Cohen's Kappa score & 0.898 \\
            \midrule
            \ \textsc{Development Corpus} & \textbf{3924} \\
            \midrule
            \ \ w entity mentions & 1493 \\
            \ \ w/o entity mentions & 2431 \\
            \midrule
            \ \textsc{Few-Shot Target Corpus} & \textbf{384} \\
            \midrule
            \ \ w entity mentions & 369 \\
            \ \ w/o entity mentions & 15 \\
            \bottomrule
        \end{tabular}
    \end{adjustbox}
    \caption{Dataset annotation and final statistics.}
    \label{table:annotation_stats}
\end{table}



\subsection{Annotation Procedure}
Both annotators reviewed and revised their work without discussion or knowledge of the other annotator's work. 
%
%
In cross-revision, the reviewing annotator only made corrections when \emph{labeling inconsistencies} were due to a lack of medical terminology comprehension. 
During revision, no changes to the original annotator's confidence level rating were made. \\
%
%


%
\textbf{Confidence Level:} The criteria for the confidence levels are annotator dependent as summarized in \cref{table:ci_examples} with examples.
PHI has been manually modified from the original data to preserve privacy while maintaining exemplary characteristics for each label entity type.\\
\textbf{Skipped Sentences:} Each annotator followed an independent set of criteria to exclude sentences from annotation, as demonstrated by examples in \cref{table:skipped_sent_samples}.
%
%
%



\begin{table}[!t]
\centering
\begin{adjustbox}{width=0.98\linewidth}
\begin{tabular}{@{}cll@{}}
\toprule
\textbf{Annotator} & \textbf{Sample sentence} & \textbf{Explanation} \\ \midrule
A &
  \begin{tabular}[c]{@{}l@{}}“Actualmente reside en \\ XXXX-Xxxx Xxxxxxxx, \\ Treballadora Social.” \\ {[}Currently resides in \\ XXXX-Xxxxx Xxxxxxxx, \\ social worker.{]}\end{tabular} &
  \begin{tabular}[c]{@{}l@{}}The underlined words \\ are grouped as a single word token. \\ From the context it’s clear \\ that `XXXX' belongs to `LOCATION' \\ and `Xxxxx Xxxxxxxx' are \\ `NAME' entities.\end{tabular} \\ \midrule
B&
  \begin{tabular}[c]{@{}l@{}}“Lmarxa. {[}sic{]}” \\ {[}March or walks{]}\end{tabular} &
  \begin{tabular}[c]{@{}l@{}}Annotator does not \\ have enough context \\ to understand this \\ token to annotate.\end{tabular}
   \\ \bottomrule
\end{tabular}
\end{adjustbox}
\caption{Examples of sentences skipped by annotators and rationales.}
\label{table:skipped_sent_samples}
\end{table}


\subsection{Annotator Disagreements and Resolution} \label{subsec:disagreement_resolution}
An attempt was made to review the 390 sentences where our annotators disagreed to find a resolution. 
Main sources of disagreement are due to: (a) annotation criteria discrepancy, (b) ambiguity between related entities, and (c) annotation errors.
After further revision to correct identified errors and clarify ambiguous annotation criteria, agreement was reached for 384 sentences while 6 sentences were left un-annotated due to insufficient context. 
Confidence levels from both annotators were left unchanged.
These sentences constitute our \textit{few-shot target corpus} in the pipeline explained in \cref{fig:transfer_strategy}.

\paragraph{Inclusion Criteria Discrepancy:} Most disagreements are related to discrepancy in the inclusion of surrounding words such as determiners, punctuation marks, and descriptive phrases. 
This is prevalent particularly in the LOCATION and PROFESSION entities. One annotator considered denoted sentences with these characteristics a lower confidence level of 4 compared to sentences without determiners or punctuation marks surrounding LOCATION tokens. 
%
%
The resolution step changed the annotations to be more consistent with the annotation guidelines described in \cref{table:anno_guidelines}.

\paragraph{Ambiguity Between Related Entities:}
Another source of disagreements in the LOCATION PHI stems from abbreviation usage and confusion with the NAME PHI. 
In instances where the syntax is ambiguous, annotators may not be able to infer correctly that certain unknown abbreviations are place names. 
Since it is common that places are named after people’s names and vice versa, lack of contextual information creates unresolvable ambiguity regarding the NAME and LOCATION entities. 
DATE and AGE also demonstrate a similar disagreement behavior.
In particular, numerical and text expressions involving ‘years’ may express age or time depending on context.
%
%
\paragraph{Annotation Errors:}
A few disagreements are due to mislabeling or erroneous omissions. 
There are fewer than 5 instances in the 390 disagreements.
%
%
%
Notable errors are associated with mislabeling proper names that resemble valid named entities. 
For instance, some assessment tools are named after people or place names e.g. Barcelona Test and Boston (Naming) Test.



\begin{table*}[!t]
    \centering
    \resizebox{\linewidth}{!}{
        \begin{tabular}{lc}
        \toprule
        \textbf{PHI} & \textbf{Fine-grained Types} \\
        \midrule
        AGE & EDAD\_SUJETO\_ASISTENCIA \\
        \midrule
        CONTACT & NUMERO\_TELEFONO, NUMERO\_FAX, CORREO\_ELECTRONICO, URL\_WEB \\
        \midrule
        DATE & FECHAS \\
        \midrule
        \multirow{4}{*}{ID} & ID\_ASEGURAMIENTO, ID\_CONTACTO\_ASISTENCIAL, NUMERO\_BENEF\_PLAN\_SALUD, \\
        & IDENTIF\_VEHICULOS\_NRSERIE\_PLACAS, IDENTIF\_DISPOSITIVOS\_NRSERIE, \\ 
        & IDENTIF\_BIOMETRICOS, ID\_SUJETO\_ASISTENCIA, ID\_TITULACION\_PERSONAL\_SANITARIO, \\
        & ID\_EMPLEO\_PERSONAL\_SANITARIO, OTRO\_NUMERO\_IDENTIF \\
        \midrule
        LOCATION & HOSPITAL, INSTITUCION, CALLE, TERRITORIO, PAIS, CENTRO\_SALUD \\
        \midrule
        NAME & NOMBRE\_SUJETO\_ASISTENCIA, NOMBRE\_PERSONAL\_SANITARIO \\
        \midrule
        PROFESSION & PROFESION \\
        \midrule
        \multirow{2}{*}{OTHER} & SEXO\_SUJETO\_ASISTENCIA, FAMILIARES\_SUJETO\_ASISTENCIA, \\ 
        & OTROS\_SUJETO\_ASISTENCIA, DIREC\_PROT\_INTERNET \\
        \bottomrule
        \end{tabular}
    }
    \caption{MEDDOCAN PHI (coarse-grained) and fine-grained types \cite{marimon2019automatic}.}
    \label{table:meddocan_types}
\end{table*}



\section{MEDDOCAN Normalization}\label{app:meddocan_normalization}
The original MEDDOCAN dataset \cite{marimon2019automatic} provides document level de-identification annotations, following 2014 i2b2/UTHealth \cite{stubbs2015annotating}, of 1000 clinical notes which are divided into 500, 250 and 250 for training, validation and testing respectively. 
It contains 29 fine-grained entity types classified into 8 coarse-grained PHI types (\cref{table:meddocan_types}). 
Compared to i2b2 (2014), MEDDOCAN has an additional OTHER category which we normalize to O in the BIO scheme, resulting in 7 coarse-grained PHI types considered in this work.
We tokenize the 500 training notes resulting in 16,299 sentences.
The conversion script is available in \textsc{T2NER}. \footnote{\url{https://github.com/suamin/T2NER/blob/master/utils/convert_i2b2style_xml_to_conll.py}}
%


\begin{table*}[!ht]
    \centering
    \begin{adjustbox}{max width =\textwidth}
    \begin{tabular}{@{}ll@{}}
    \toprule
    \textbf{PHI} & \textbf{Criteria} \\ \midrule
          AGE
          & 
          \begin{tabular}[c]{@{}l@{}}Annotate only the numerical part of the expression; include both numerical and word expressions of age (e.g. 36 or thirty-six ). \\ 
          Include the words ‘years’, ‘months’, and ‘days’ when they express age. \\ 
          Include expressions that describe an age group e.g. ‘adolescent’, ‘recently born’, ‘new born’. \\
          Include punctuation associated with age, including separate tokens, e.g. in his/her 30’s. 
          \end{tabular} \\ \midrule
          CONTACT
          & 
          \begin{tabular}[c]{@{}l@{}}All forms of contact information, e.g. pager, phone numbers, e-mail address. \\ 
          Physical or mailing address is annotated as ‘Location’ \\
          Include punctuation and symbols that occur with contact information, e.g. include all tokens in ‘(123) 456-789’. 
          \end{tabular} \\ \midrule
          DATE
          & 
          \begin{tabular}[c]{@{}l@{}}Include days of the week and months. \\ 
          Include punctuation in all formats. \\
          Include the word ‘year’ and ‘month’ that are part of a date-time expression, e.g. ‘the year 2000’. \\
          Include prepositions that are part of a date-time expression, e.g. include the word ‘of’ in ‘5th of May’. 
          \end{tabular} \\ \midrule
          ID
          & 
          \begin{tabular}[c]{@{}l@{}}Include all identification numbers such as Medical Record Number (MRN), Social Security Number (SSN), Document ID, device lot number, etc. \\ 
          Include any alpha-numeric expressions appearing in the beginning of the document or next to a name that's not formatted as a date.  \\ 
          When separated by punctuation, annotate all parts of the expression including punctuation, e.g. include all tokens in ‘12-34-5678’. \\
          Exclude the ID descriptive words and associated punctuation, e.g. exclude ‘MRN’ and ‘:’ in ‘MRN: 1234567’. 
          \end{tabular} \\ \midrule
          LOCATION
          & 
          \begin{tabular}[c]{@{}l@{}}Include all place names and all parts of an address: street name, city, state, county, province, region, and country.  \\ 
          Include punctuation in address. \\ 
          Include Zip/postal codes. \\
          Include organization names. \\
          Include words that specify location when they appear as part of a ‘Location’ entity,  e.g. include the word ‘Center’ in ‘Social Security Center’.
          \end{tabular} \\ \midrule
          NAME
          & 
          \begin{tabular}[c]{@{}l@{}}Include only the person's names. \\ 
          Include punctuation between first and last names when present \\ 
          Exclude titles and salutations. 
          \end{tabular} \\ \midrule
          PROFESSION
          & 
          \begin{tabular}[c]{@{}l@{}}Include all professional titles, e.g. annotate ‘MD’ in the phrase ‘X works as an MD’. \\
          Exclude professional titles in name suffixes, e.g. exclude ‘MD’ in the phrase ‘Dr. X Y, MD’. \\ 
          Include professional and occupational descriptions, e.g. annotate ‘carpentry in the phrase ‘X works in carpentry’. \\
          Annotate the entire expression describing a profession, e.g. annotate all tokens in a phrase such as ‘worker in a cafeteria’. \\
          Exclude workplace names; annotate workplace names as ‘Location’ instead. 
          \end{tabular} \\ \bottomrule
    \end{tabular}
    \end{adjustbox}
    \caption{Adjusted annotation guidelines with examples for each PHI type.}
    \label{table:anno_guidelines}
\end{table*}



\begin{table*}[!t]
\centering
\begin{adjustbox}{max width =\textwidth}
\begin{tabular}{@{}cllll@{}}
\toprule
\multicolumn{1}{l}{\textbf{Level}} & \textbf{Annotator} & \textbf{Criteria} & \textbf{Sample Sentence} & \textbf{Explanation} \\ \midrule
      \multirow{2}{*}[-2em]{1} 
      & 
      A 
      & 
      \begin{tabular}[c]{@{}l@{}}Annotator is unable to assign \\ 
      labels due to insufficient \\
      contextual information from \\
      the given sentence. 
      \end{tabular}
      &
      \begin{tabular}[c]{@{}l@{}}``PASe." \\ {[PASe.]} or [ENTer.] \end{tabular}
      &
      \begin{tabular}[c]{@{}l@{}}The token may be an unknown \\
      acronym or an oddly typed \\
      imperative form of the verb \\
      ``to enter". Insufficient context.
      \end{tabular}
      \\ 
      \cmidrule(l){2-5}
      &
      B
      &
      \begin{tabular}[c]{@{}l@{}}Annotator is unable to assign \\ 
      labels due to\\ 
      lack of comprehension. 
      \end{tabular}
      &
      \begin{tabular}[c]{@{}l@{}}``Allitat, en DDLL.” \\ {[n/a]} 
      \end{tabular}
      &
      \begin{tabular}[c]{@{}l@{}}Annotator did not understand \\ 
      this sentence in Catalan \\
      sufficiently to annotate. 
      \end{tabular} \\ 
      \midrule
      \multirow{2}{*}[-3em]{2} 
      & 
      A 
      & \begin{tabular}[c]{@{}l@{}}Annotator is unsure \\
      about the assigned labels \\ 
      due to contextual ambiguity. 
      \end{tabular}
      &
      \begin{tabular}[c]{@{}l@{}}``“50 años.” \\ {[50 years]} \end{tabular}
      &
      \begin{tabular}[c]{@{}l@{}} Without any surrounding context, \\
      the years can be ‘AGE’ \\
      or a temporal expression; \\
      annotator thinks it’s most likely to be AGE, \\
      but does not feel confident enough \\
      to make a determination. 
      \end{tabular}
      \\ \cmidrule(l){2-5}
      &
      B
      &
      \begin{tabular}[c]{@{}l@{}}Annotator is unsure about \\
      the assigned labels due \\
      to lack of medical knowledge \\
      or terminology. 
      \end{tabular}
      &
      \begin{tabular}[c]{@{}l@{}}``Urocultiu {[sic]} 13.01: + \\ per \textit{A. baumanii} multiR.” \\
      Urine Culture 13.01: + \\ for MDR \\ \textit{A. baumanii} 
      \end{tabular}
      &
      \begin{tabular}[c]{@{}l@{}}Annotator omitted this sentence \\
      due to uncertainty about \\ the word \textit{A. baumanii}, whether it \\
      could be a NAME or a \\ non-labelled entity. 
      \end{tabular} \\ 
      \midrule
      \multirow{2}{*}[-2em]{3} 
      & 
      A 
      & 
      \begin{tabular}[c]{@{}l@{}}Annotator is confident about \\
      the labels, but some context \\
      may be missing that could \\
      change the entity labels. 
      \end{tabular}
      &
      \begin{tabular}[c]{@{}l@{}}``712345678)." \end{tabular}
      &
      \begin{tabular}[c]{@{}l@{}}This is likely a phone number \\ CONTACT, but may \\
      also be an ID entity. 
      \end{tabular}
      \\ \cmidrule(l){2-5}
      &
      B
      &
      \begin{tabular}[c]{@{}l@{}}Annotator is confident about the sentence \\
      in general, but has some doubt \\
      due to presumed lack of \\
      specialized knowledge. 
      \end{tabular}
      &
      \begin{tabular}[c]{@{}l@{}}``hipoTA {[sic]} asintomática."  \\
      {[Asymptomatic hypotension.]}
      \end{tabular}
      &
      \begin{tabular}[c]{@{}l@{}}Annotator did not specify any \\
      label but was unsure whether \\
      there was a labelled entity \\
      or not.  
      \end{tabular} \\ \midrule
      \multirow{2}{*}[-2em]{4} & A & \begin{tabular}[c]{@{}l@{}}Annotator is confident \\
      about the labels, but the sentence may \\
      have some inconsistencies with \\
      the gold standard sentences.
      \end{tabular}
      &
      \begin{tabular}[c]{@{}l@{}}"El marido la vió y llamó \\
      a la ambulancia e ingresó \\
      en el hospital de Xxxxxxx." \\ 
      {[}The spouse saw her and \\
      called the ambulance and \\
      she was admitted to Xxxxxxx \\
      hospital.{]}
 \end{tabular}
      &
      \begin{tabular}[c]{@{}l@{}}Annotator was unsure whether \\
      to only annotate `Xxxxxxx' or \\ 
      `hospital de Xxxxxxx' or \\
      `el hospital de Xxxxxxx' as LOCATION 
      \end{tabular}
      \\ \cmidrule(l){2-5}
      &
      B
      &
      \begin{tabular}[c]{@{}l@{}}Annotator is confident about \\
      the labels, but the sentence may \\
      have some inconsistencies with \\
      the gold standard sample sentences. 
      \end{tabular}
      &
      \begin{tabular}[c]{@{}l@{}}``Torna d'Oftalmologia de \\
      Xxx Xxxx ( Dra. {[sic]}" \\ 
      {[}Returns from Xxx Xxxx \\
      Ophthalmology (Dra. {]} 
      \end{tabular}
      &
      \begin{tabular}[c]{@{}l@{}}Annotation unsure whether or \\
      not to include Ophthalmology \\
      as part of `LOCATION' 
      \end{tabular} \\ \midrule
      \multirow{2}{*}[-5em]{5} & A & \begin{tabular}[c]{@{}l@{}}Annotator is confident \\
      and there’s no ambiguity regarding \\
      name entities of the labels. \\
      This could mean that the sentences \\
      have no entities to be annotated \\
      or that all the entities needing \\
      annotations are consistent with \\
      the gold standard sample sentences. 
      \end{tabular}
      &
      \begin{tabular}[c]{@{}l@{}}``Cito a control el próximo \\
      25.12.20 y doy pautas \\
      a la esposa." \\
      {[}I make a follow-up \\
      appointment for the upcoming date \\
      25.12.20 and I give the \\
      prescription to the wife.{]} 
 \end{tabular}
      &
      \begin{tabular}[c]{@{}l@{}}It’s clear that `25.12.20' is a \\
      DATE PHI. 
      \end{tabular}
      \\ \cmidrule(l){2-5}
      &
      B
      &
      \begin{tabular}[c]{@{}l@{}}It is clear to the annotator that \\
      the sentence has no entities to \\
      be annotated or that the entities \\
      are consistent with gold standard \\
      annotation. This could be either \\
      apparent at first glance or because \\
      the sentence has been seen several \\
      times before, which increases the \\
      annotator's confidence regarding \\
      the assigned label(s). 
      \end{tabular}
      &
      \begin{tabular}[c]{@{}l@{}}``Cito a control el próximo \\
      25.12.20 y doy pautas \\
      a la esposa." \\
      {[}I make a follow-up \\
      appointment for the upcoming date \\
      25.12.20 and I give the \\
      prescription to the wife.{]} 
      \end{tabular}
      &
      \begin{tabular}[c]{@{}l@{}}It’s clear that `25.12.20' is a \\
      DATE PHI. 
      \end{tabular} \\ 
      \bottomrule
\end{tabular}
\end{adjustbox}
\caption{Confidence level criteria and examples as reported by the two annotators. In instances where PHI entities are utilized in the examples, we replaced the characters with generic alphanumeric characters or with fictitious information (while maintaining the same PHI type).}
\label{table:ci_examples}
\end{table*}


\end{document}